\documentclass[lettersize,journal]{IEEEtran}
\usepackage{amsmath,amsfonts}
\usepackage{amsthm}
\usepackage{amssymb}
\usepackage{algorithmic}
\usepackage{algorithm}
\usepackage{array}
\usepackage[caption=false,font=normalsize,labelfont=sf,textfont=sf]{subfig}
\usepackage{textcomp}
\usepackage{stfloats}
\usepackage{url}
\usepackage{verbatim}
\usepackage{graphicx}
\usepackage{cite}
\usepackage{xspace}
\usepackage[dvipsnames]{xcolor}
\usepackage{todonotes}

\newtheorem{theorem}{Theorem}
\newtheorem{definition}{Definition}
\newtheorem{proposition}{Proposition}
\newtheorem{corollary}{Corollary}
\newtheorem{lemma}{Lemma}
\newtheorem{example}{Example}

\newcommand{\bdfn}{\begin{definition}}
\newcommand{\edfn}{\end{definition}}
\newcommand{\bthm}{\begin{theorem}}
\newcommand{\ethm}{\end{theorem}}
\newcommand{\bprop}{\begin{proposition}}
\newcommand{\eprop}{\end{proposition}}
\newcommand{\blem}{\begin{lemma}}
\newcommand{\elem}{\end{lemma}}
\newcommand{\bcor}{\begin{corollary}}
\newcommand{\ecor}{\end{corollary}}
\newcommand{\bex}{\begin{example}}
\newcommand{\eex}{\end{example}}
\newcommand{\bobs}{\begin{observation}}
\newcommand{\eobs}{\end{observation}}

\newcommand{\refun}{Function-Representation\xspace}
\newcommand{\refuns}{Function-Representations\xspace}

\newcommand{\calH}{{\cal H}}

\begin{document}

\title{The \refun Model of Computation}

\author{\IEEEauthorblockN{Alfredo Ibias\IEEEauthorrefmark{1}, Hector Antona, Guillem Ramirez-Miranda and Enric Guinovart}\\
\IEEEauthorblockA{Avatar Cognition\\
Barcelona, Spain\\
Email: \{alfredo\IEEEauthorrefmark{1}, hector, guillem, enric\}@avatarcognition.com}\\
\and
\IEEEauthorblockN{Eduard Alarcon}\\
\IEEEauthorblockA{Universitat Politècnica de Catalunya - BarcelonaTech\\
Barcelona, Spain\\
Email: eduard.alarcon@upc.edu\\
}
\thanks{Manuscript received MMMM DD, 20YY; revised MMMM DD, 20YY.}
\IEEEauthorrefmark{1} Corresponding author}

\markboth{IEEE Transactions on Neural Networks and Learning Systems,~Vol.~XX, No.~X, MMMM~20YY}%
{Ibias \MakeLowercase{\textit{et al.}}: The \refun Model of Computation}

\IEEEpubid{0000--0000/00\$00.00~\copyright~2021 IEEE}

\maketitle

\begin{abstract}
Cognitive Architectures are the forefront of the research into developing an artificial cognition. However, they approach the problem from a separated memory and program model of computation. This model of computation poses a fundamental problem: the knowledge retrieval heuristic. In this paper we propose to solve this problem by using a novel model of computation, one where memory and program are merged: the \refun. This model of computation involves defining a generic \refun and instantiating multiple instances of it. In this paper we explore the potential of this novel model of computation through mathematical definitions and proofs. We also explore the kind of functions a \refun can implement, and present different ways to organise multiple instances of a \refun.
\end{abstract}


\section{Introduction}\label{sec:intro}
\IEEEPARstart{A}{} Cognitive Architecture is a model of how the mind is thought to work, that has an implementation in the form of computer instructions. Thus, the end goal of any Cognitive Architecture is to be able to produce cognition inside a computer. To approach this goal, most Cognitive Architectures are based on the model of computation our computers use, where there is a \emph{memory} that stores the knowledge, and separately there is a set of \emph{programs} that update, manipulate and work with the memory. This memory usually stores the knowledge in the form of \emph{representations}, that are either prototypes or exemplars of the knowledge that is being represented. This kind of Cognitive Architecture is what has been termed a symbolic Cognitive Architecture, whose more famous examples are SOAR~\cite{Laird08, Laird12} and ACT-R~\cite{VanLehn86}. An alternative model of computation is the connectionist approach, also known as parallel distributed processing, in which there is a set of processing units from which the functional behaviour of the system emerges. However, in this approach usually the processing units also separate memory from program, by having a smaller memory full of partial representations while the manipulation functions are set apart. A famous example of the connectionist approach is Copycat~\cite{Mitchell90, Burns95}, although Artificial Neural Networks could also fall in this category if considered as a Cognitive Architecture.

The main problem of using separated memory and program is that, when needing to retrieve knowledge to work with, there is no clear integrated solution that will provide the right representations from memory. Moreover, there is a need for complex heuristics that, given the context, try to estimate which are the representations that are needed to be retrieved from the memory. This poses a huge problem, as it requires looking for a general algorithm that can, in front of any eventuality, find the correct representations in a constantly evolving memory. This is in fact like looking for the algorithm that solves any problem, as any knowledge retrieval algorithm has to know the solution for any problem that cognition aims to solve. This has not impeded each Cognitive Architecture to propose its own heuristics to perform this task, with more recent developments even proposing using a collection of heuristics instead of only one~\cite{Lieto16, pc17}. However, none of those Cognitive Architectures have been able to produce cognition in a computer.

In this paper we address this problem from a different optic. Instead of looking for such heuristic, we go back to the fundamental model of computation of the Cognitive Architecture and propose a critical change. Instead of developing Cognitive Architectures that separate memory and program, we propose to build Cognitive Architectures where the representation is the function. In this new setting, we will have a set of representations (memory) that at the same time will constitute the function (program) of the Cognitive Architecture. In that regard, we would need a connectionist approach, where each processing unit is a representation, and the function is determined by such representation. Then, from the connections between different processing units the cognitive behaviour should emerge.
\IEEEpubidadjcol

This approach is not actually novel, as recent Kolmogorov-Arnold Networks~\cite{ii24} and Artificial Neural Networks actually follow it, although at different scales of complexity and abstraction. In theses cases, each neuron is a \refun where its weights are the representation and its function is the processing of the input with its weights (and the activation function) to produce an output that is passed onto the next neuron. In this paper we aim to provide a theoretical foundation to this approach, proving mathematically some of their properties and exploring how it solves the knowledge retrieval problem present in current Cognitive Architectures. However, although we pose Artificial Neural Networks as an example of this approach, and we will use them later to show empirical results, we will not limit ourselves to them. Moreover, we will try to accommodate as many possible Cognitive Architectures as possible, as long as they follow this approach in which the representation is the function.

In this paper we define \refun as a parametrised function, where the parameter values are the representation. In that regard, when connecting multiple \refuns, we consider that we are connecting multiple instances of the same parametrised function. Thus, to simplify the description, we will use the terms \refuns and instances of a \refun indistinguishable, to refer to multiple instances of the same parametrised function. In this paper we also prove how this new approach can store knowledge, how connecting multiple \refuns can produce emergence, and explore some characteristics the parametrised functions should have in order to produce a Cognitive Architecture. Finally, we prove mathematically our claims, explore different use cases of our proposal, and analyse some limitations of our work.

The rest of the paper is organised as follows. In Section~\ref{sec:work} we briefly present related work about Cognitive Architectures. In Section~\ref{sec:fram} we present our model of computation. In Section~\ref{sec:proofs} we present the mathematical assumptions and proofs that support our work. In Section~\ref{sec:func} we delve a bit more into which properties define which functions. In Section~\ref{sec:cases} we present some case scenarios to show the potential of our model of computation. In Section~\ref{sec:disc} we discuss some limitations of our model of computation. And finally, in Section~\ref{sec:conc} we present the conclusions of our work.

\section{Related Work}\label{sec:work}
Cognitive Architectures can be roughly divided into three types, based on their knowledge processing patterns~\cite{yww18}: symbolic, emergent, and hybrid. The symbolic Cognitive Architectures store their representations as symbols and usually operate with formal logic. The emergent Cognitive Architectures focus on reproducing the human brain, and usually adopt a hierarchical structure. Finally, hybrid Cognitive Architectures try to find a middle ground between both, having hierarchical structures, behaviour emergence and symbolic processing.

The symbolic Cognitive Architectures build over the idea that our brains work with symbols and abstractions. Thus, all of them represent the environment as symbols and store them in a consistent knowledge base. To operate with the symbols, they tend to use formal logic, although more modern proposals try to exploit the advantages of fuzzy logic. Thus, due to their properties, they are mainly used for system control problems, like the ones found in robotics. Inside this group, there are some Cognitive Architectures designed to solve concrete problems, like the control of military unmanned vehicles (4D/RCS~\cite{Albus00,abchss06}), the control of robots (ADAPT~\cite{bll04, bll04b}, CARACaS~\cite{haht11}, REAPER~\cite{mmadfjj+01}), the processing of images (DSO-CA~\cite{pttan12,nxct12}), the support of decision-making for crisis response (NARS~\cite{Wang13,swxw15}), and the interaction with humans (DIARC~\cite{skms06}). There is also a group of Cognitive Architectures that aim to solve the general intelligence problem, like \emph{Adaptive Control of Thought-Rational} (ACT-R)~\cite{VanLehn86}, \emph{Belief-Desire-Intention} (DBI)~\cite{gpptw98}, \emph{Reflective Evolutionary Mind} (REM)~\cite{mg01} and \emph{State, Operator, and Result} (SOAR)~\cite{Laird08, Laird12}.

The emergent Cognitive Architectures try to ``reproduce'' the process of human cognition from the bottom up. To that end, they adopt a hierarchical structure in which the lower levels deal with perception while the higher levels correspond to cognitive processes. In that regard, the lower levels tend to run concurrently while the higher levels solve the possible conflicts of consistency and choose the behaviour to perform. This setup allows them to deal with uncertainty conditions. Inside this group we also have the differentiation between Cognitive Architectures designed to solve particular problems, like the control of robots (ASMO~\cite{njw10}, DAC~\cite{mbv12}, MDB~\cite{bfpd06}), and Cognitive Architectures that aim to solve the general intelligence problem, like \emph{Adaptive, Reflective Cognition in an Attention-Driven Integrated Architecture} (ARCADIA)~\cite{bb15}, \emph{Learning Intelligent Distribution Agent} (LIDA)~\cite{fmsfd+16}, and \emph{Sensory-Motor, Episodic Memory and Learning, and Central Executive} (SEMLC)~\cite{js13}. Additionally, there is an additional set of Cognitive Architectures that are based on computational neuroscience, like \emph{Brain-Based Device} (BBD)~\cite{edelman07} and \emph{Hierarchical Temporal Memory} (HTM)~\cite{ha15}, that aim to simulate the brain's function.

Finally, the hybrid Cognitive Architectures take the best parts of each alternative, and try to blend them into a consistent architecture able to produce cognition. To that end, they usually adopt a hierarchical structure and contain both behaviour emergence from the bottom up together with top-down direct symbolic processing. Inside this group there are mainly Cognitive Architectures that aim to solve the general intelligence problem, like \emph{Chunk Hierarchy Retrieval Structures} (CHREST)~\cite{lklg14}, \emph{Connectionist Learning With Adaptive Rule Induction Online} (CLARION)~\cite{spm99}, and \emph{Synthesis of ACT-R and Leabra} (SAL)~\cite{jloa08}.

What all these Cognitive Architectures have in common is that they tend to use the same terms to refer to its different parts~\cite{llr17}. Short Term Memory (STM), Long Term Memory (LTM) and Working Memory (WM) are the main kinds of memory explored, all in a model of computation in which the memory is independent of the program. Perception and actuation are the main methods to interact with the world, while learning and attention are the main methods to update the memories and the reasoning functions. However, all these Cognitive Architectures follow the separated memory and program model of computation presented in Section~\ref{sec:intro}, and thus all of them are limited by their knowledge retrieval heuristics. The only approach to human intelligence that has avoided this problem is the Artificial Neural Networks. However, those have never been presented as a Cognitive Architecture, but instead mainly as an optimisation method. Thus, they have not tried to build the elements that all Cognitive Architectures aim to develop.

On a different note, there is also related work about mixing memory and program into a single computational unit. Mainly, this work has been developed in the computer architecture field, where the main proposal is \emph{in-memory processing}~\cite{gbkgm19}. In this approach, the physical memory has the capability of performing basic operations such as matrix-vector multiplication, not enough for Turing complete computation, but sufficient for their purpose. Its main uses have been to develop \emph{Tensor Processing Units}~\cite{klr19}, for processing Artificial Neural Networks more efficiently, and for reducing the computational cost of \emph{Hyperdimensional computing}~\cite{rkr16}.

\section{The \refun Model of Computation}\label{sec:fram}
In this section we aim to present the broad aspect of our model of computation, and we will provide some examples of implementations to illustrate it.

Our model of computation revolves around the idea that the representation operates itself also as the function. Thus, its main component is this \refun. A \refun must have a special form, as not any representation can be used as a function. For example, a set of weights (like in an Artificial Neural Network) is a good \refun because those weights can process any input and produce an output. However, a word or symbol is not a valid \refun as some inputs cannot be processed with them.

In a more general sense, a representation that can be used as a function needs to fulfil a fundamental requisite: that it can be used to process an input and produce an output. Thus, its data type should be defined according to its function. In the end, a representation that is also a function can be distilled to a pair of data type and function, where the data type defines the kind of values the representation will store, and the function will use those values to process any input to produce an output. An example would be an image (this is, a vector of numbers) and an average function, that for an input image would produce an output image that is the average of the representation image and the input image. In a more formal way, a \refun is a function parameterised by a set of values of a certain data type.

The other aspect of our model of computation is the need for connectionism. A processing unit by itself can only contain a \refun and produce an output for each input it receives, but that is not enough to build a Cognitive Architecture. Thus, there is a fundamental need for connecting multiple processing units with one another. The specific configuration of the connections is not defined, as each kind of \refun will have its own reasons to connect some processing units between themselves and not others. For example, in Artificial Neural Networks, the connections are done by layers, but not between the neurons in the same layer.

The only requirement for the connections is that they provide the output of a \refun to another one, thus using the output of a \refun as input for the following one. This in turn creates a hierarchical structure that produces a higher level behaviour that emerges from the individual behaviour of each \refun. In the previous image example, connecting one \refun with the following one in turn produces a pattern matching algorithm.

Finally, there is the question of learning. A \refun needs to be learned, and new \refuns can be created if needed. However, these mechanisms depend a lot on the kind of \refun that is being built. For example, Artificial Neural Networks use backpropagation to learn based upon an error measure, and do not create new \refuns, but in the previous image example, we may not want to update the \refun with the processed input, but instead create new \refuns at some point to expand the stored knowledge and improve the pattern matching capability.

Eventually, the goal is to follow a novel model of computation in which individual \refuns (with very simple computational mechanisms) are the fundamental units of computation, and the complex behaviours arise from the interactions between those \refuns. To that end, it is critical to avoid global algorithms that organise or modify the \refuns, but instead, develop mechanisms inside the \refuns that allow them to self-organise and self-modify. For example, in Artificial Neural Networks, even backpropagation (commonly thought to be a global learning algorithm) can be translated into a function of the own neurons, fulfilling the \refun definition. We argue that this fact is key to explain why the Artificial Neural Networks have been able to obtain the outstanding results they have obtained, in so many different scenarios, without needing to change the fundamental algorithm.

However, we are aware that changing the computational paradigm is difficult. For example, Artificial Neural Networks have been developed following this model of computation unconsciously, applying the current separated memory and program model of computation and obtaining an algorithm that is equivalent in both models of computation. However, learning to program in this novel model of computation is fundamental to develop Cognitive Architectures able to eventually result in emergence of cognition.

\section{Proofs}\label{sec:proofs}
In this section we will explore some properties of the proposed model of computation. Specifically, we will explore how \refuns contain knowledge, how that knowledge can be accessed without heuristics, and how higher-level behaviour emerges from its connectionism.

To begin with this, we need to find a definition for knowledge. As this is more of a philosophical question that we do not aim to solve in this paper, we decided to use a more pragmatic approach and define knowledge based on its utility, that is, in its capacity to transform data. Thus, we will define knowledge as a function that can be used to transform any input set into information, as defined by Shannon's Information Theory~\cite{sha48}. This theory defines information as the amount of uncertainty involved in the value of a random variable, and measures such information using entropy:
\bdfn\label{dfn:entropy}
Let $A$ be a finite set and $\xi_A$ be a random variable over $A$. We denote by $\sigma_{\xi_A}$ the probability distribution induced by $\xi_A$. The \emph{entropy} of the random variable $\xi_A$, denoted by $\calH(\xi_A)$,  is defined as:
$$\calH(\xi_A)=-\sum_{a\in A} \sigma_{\xi_A}(a)\cdot \log_2 (\sigma_{\xi_A}(a))$$

The set $A$ is said to contain information if and only if $\calH(\xi_A) > 0$. If $\calH(\xi_A) = 0$ then the set $A$ is said to not contain information.
\edfn
Now, with this definition, let us define how a function over a set transforms random variables.
\bthm\label{thm:transform}
Given two sets $I$ and $O$ and a function $f: I \to O$. Let $\xi_I$ be a random variable over $I$ such that it induces a uniform probability distribution $\sigma_{\xi_I}$, then the function $f$ transforms $\xi_I$ into a random variable $\xi_O$ over the set $O$ such that it induces another probability distribution $\sigma_{\xi_O}$.
\ethm
\begin{proof}
    Given the random variable $\xi_I$, as it produces an uniform probability distribution $\sigma_{\xi_I}$ over $I$, it implies that one can pick each element of $I$ with the same probability. If we then process those elements of $I$ with $f$, we get a set of elements of $O$, each one with its own probability based on how many elements of $I$ are transformed via $f$ into such element of $O$. Thus, this produces a probability distribution $\sigma_{\xi_O}$ that has to come from a random variable $\xi_O$ over $O$. Therefore, $f$ transforms $\xi_I$ into $\xi_O$.
\end{proof}
This theorem is important to show how, under the assumption that the inputs of a function are provided uniformly, this function produces a random variable over the output set, what we will use now to define the information produced by such function.

\bthm\label{thm:info}
Given two sets $I$ and $O$ and a function $f: I \to O$. Let $\xi_I$ be a random variable over $I$ such that it induces an uniform probability distribution $\sigma_{\xi_I}$, and let the function $f$ transform $\xi_I$ into a random variable $\xi_O$ over the set $O$ such that it induces another probability distribution $\sigma_{\xi_O}$. Then, the output set $O$ contains information via $f$ if and only if the entropy of $\xi_O$ is greater than $0$.
\ethm
\begin{proof}
    By Definition~\ref{dfn:entropy}, as $\xi_O$ is a random variable over $O$ (produced from the random variable $\xi_I$ over $I$ via $f$), then it is said to contain information if and only if $\calH(\xi_O) > 0$. Thus, by extension, $O$ contains information via $f$ due to $f$ being the function producing $\xi_O$.
\end{proof}

Now, let us define a function as knowledge (based on the restricted definition of knowledge considered in this work) if and only if its output set has more than one value.
\bthm\label{thm:knowledge}
Given two sets $I$ and $O$ and a function $f: I \to O$, then  $f$ is knowledge if it transform inputs $i \in I$ into outputs $o \in O$ and the size of the output set $O$ is greater than $1$. That is, $|O| = |\{o | f(i) = o,\ with\ i \in I, o \in O\}| > 1$.
\ethm
\begin{proof}
    Let us start assuming a uniform distribution over $I$. By Theorem~\ref{thm:transform} $f$ produces a random variable $\xi_O$ over the set $O$. As the set $O$ is defined by the possible values that $f$ can produce from the set $I$, then $|O| = 1$ if and only if all the elements of $I$ are transformed into the same element. By Theorem~\ref{thm:info} the output set $O$ contains information via $f$ if and only if the entropy of $\xi_O$ is greater than $0$. And that only happens when $|O| > 1$ by Definition~\ref{dfn:entropy}.
\end{proof}
This knowledge definition allows us to say that any function that transforms inputs into outputs is knowledge, as long as the output set size is greater than $1$. That implies that an Artificial Neural Network neuron whose set of weights is all zeros is not knowledge, because they transform any input into the same output value: $0$.

Now, we need to define a \refun with its data type and its function.
\bdfn\label{def:refun}
Given a data type $\mathcal{D}$ with a set of possible values $D$, and a function $f: D \times D \to D$, then a \refun is the function $f$ parameterised by a value (or set of values) $v\in D$.
\edfn
This basic definition allows for the data type $\mathcal{D}$ to be any kind of data type, as long as a function $f$ exists.

With the definitions of knowledge and \refun, we can now prove that a \refun contains knowledge.
\bthm\label{thm:know}
Given a \refun $r$ with data type $\mathcal{D}$ (with a set of possible values $D$) and function $f: D \times D \to D$, then $r$ contains knowledge if and only if $f$ is not a constant function.
\ethm
\begin{proof}
    First, let us prove that if $f$ is not a constant function, then $r$ contains knowledge. Let us assume that $r$ does not contain knowledge, then, by Theorem~\ref{thm:knowledge}, the size of the output set should be $1$. That is, $|O| = |\{o | f(i, v) = o,\ with\ i \in D, v \in D, o \in D\}| = 1$. Now, as $f$ is not a constant function, then $\exists c_1, c_2 \in D, i_1, i_2\in D | f(i_1, v) = c_1\ and\ f(i_2, v) = c_2$. Thus, $|O| > 1$ and therefore $r$ contains knowledge.

    Now, let us prove that if $r$ contains knowledge, then $f$ is not a constant function. Let us assume that $f$ is a constant function. The fact is that, as $f$ is a constant function, then it always returns the same value for any input. Thus, $\exists c\in D | f(i, v) = c \forall i\in D, v\in D$ for a specific $c \in D$. Now, as $r$ contains knowledge, $|\{o | f(i, v) = o,\ with\ i \in D, v \in D, o \in D\}| > 1$. This means that $f$ has an output set with a size higher than $1$, thus $f$ can not be a constant function.
\end{proof}
This result is critical, as this is the base over which the rest of proofs are built. This also implies that any kind of \refun has the potential to contain knowledge.

Now, let us explore how we can access the knowledge in a \refun without using complex heuristics. This stems from the fact that the representations include their own functions, thus, for any input, they process them with their function and retrieve the output, that is, the input has been processed with the knowledge that the \refun contains.

\blem\label{lem:know}
Given a \refun $r$ with data type $\mathcal{D}$ (with a set of possible values $D$), values $v\in D$, and non-constant function $f: D \times D \to D$, and given an input $i\in D$, then the output $o = f(i, v)$ has been processed with the knowledge contained in $r$.
\elem
\begin{proof}
    As $r$ is composed of the function $f$ and the set of values $v$ by Definition~\ref{def:refun}, and $f$ is knowledge by Theorem~\ref{thm:know} because it is not constant with the values $v$. Then, $o=f(i, v)$ has been processed by the knowledge stored in $f$ and $v$, that is, the knowledge contained in $r$. It is actually a transformation of $i$ by the knowledge of $f$ and $v$.
\end{proof}
This shows both how the knowledge is easily accessed, as well as how limited is the use of such knowledge, as it can only be used to perform the transformation defined by function $f$. This in fact is a constraint that limits the use of a single \refun. However, this shortcoming can be overcome by connecting multiple \refuns among them and therefore producing emergent behaviour.

To prove the emergence of this global behaviour, first we need to define what is a connection between \refuns.
\bdfn
Given two \refuns $r_1$ and $r_2$ both with data type $\mathcal{D}$ (with a set of possible values $D$), with values $v_1\in D$ and $v_2\in D$ respectively, and with functions $f_1: D \times D \to D$ and $f_2: D \times D \to D$ respectively, a connection forms a pipeline in which, for any input $i\in D$, $r_1$ is applied first producing an output $o_1\in D$ and then $r_2$ is applied over such output to produce a final output $o\in D$. That is, a connection is the function $f^{\circ}(i) = f_2(f_1(i, v_1), v_2) = o$.
\edfn
With this definition, we can observe how connections work to compose the individual functions of each \refun into a global, larger, composite function. Here it is important to notice that, for example, in the case of Artificial Neural Networks, the function is not only the multiplication of the weights by the input and the sum of the results, but also the application of the activation function.

Finally, we just need to explore what is needed for the function result of connecting multiple \refuns to produce an emergent behaviour. In that sense, we will consider that there is emergent behaviour if none of the individual \refuns could have produced such function.
\bdfn\label{def:emerg}
A connection $f^{\circ}$ of different parameterisations of function $f$ with parameters $v_1, v_2$ produces \emph{emergent behaviour} if and only if $f^{\circ}=f(v_2)\circ f(v_1)$ cannot be simplified to a parameterisation of $f$.
\edfn
We are aware that this definition of emergent behaviour does not cover all possible kinds of emergent behaviour, but it covers at least one kind of them, and that is enough for us to prove that some kind of emergent behaviour is produced.
\bthm
Given two \refuns $r_1$ and $r_2$ both with data type $\mathcal{D}$ (with a set of possible values $D$), with values $v_1\in D$ and $v_2\in D$ respectively, and with function $f: D \times D \to D$, and a connection between them that produces the function $f^{\circ}(i) = f(f(i, v_1), v_2)$, then the function $f^{\circ}$ produces emergent behaviour if and only if $f$ is not a linear function.
\ethm
\begin{proof}
    Let us start by proving that if $f^{\circ}$ produces emergent behaviour, then $f$ should not be a linear function. Let us assume that $f$ is a linear function, then its connection can be simplified to a linear function. Now, as $f^{\circ}$ produces emergent behaviour, then $f^{\circ}=f\circ f$ cannot be simplified to a linear function by Definition~\ref{def:emerg}, and thus $f$ should not be linear.

    Let us now prove that if $f$ is not linear, then $f^{\circ}$ produces emergent behaviour. Let us assume that $f^{\circ}$ does not produce emergent behaviour, then by Definition~\ref{def:emerg} $f^{\circ}=f\circ f$ can be simplified to a linear function. As $f$ is not a linear function, the connection $f\circ f$ cannot be simplified to a linear function due to the non linearity present in it. Thus, $f^{\circ}$ should produce emergent behaviour.
\end{proof}
This result is fundamental not only because it shows how the connections can produce functions that can not be achieved using a single \refun, but also it is important because it forces the function of the \refun to be non linear, as any linear function would not produce an emergent behaviour (at least of the kind we have defined). To strengthen this conclusion, let us explore what occurs when the functions of the \refuns are linear. To achieve this, we first need to define what a self-similar function is.
\bdfn\label{def:self_similar}
Given a parameterised function $f: D \times D \to D$, and a function $f^{\circ} = f \circ f$, then $f^{\circ}$ is self-similar if and only if $f^{\circ} \equiv f$.
\edfn
This definition will be useful to generalise our findings, as what we are going to explore is not limited to linear functions. However, we also need to prove that the connection of linear functions is self-similar.
\bthm\label{thm:self-similar}
Given a parameterised linear function $f: D \times D \to D$, and a function $f^{\circ}=f \circ f$ that is the connection of two parameterisations of $f$, then $f$ is self-similar.
\ethm
\begin{proof}
    Let us get two parameterisations $v_1\in D$ and $v_2\in D$ of $f$ such that $f^{\circ} = f(v_2) \circ f(v_1)$. Then $f^{\circ}(i) = f(f(i, v_1), v_2)$, but as $f$ is a linear function, then $\exists v' \in D | f(f(i, v_1), v_2) = f(i, v') \forall i \in D$. This makes $f^{\circ}(i) \equiv f(i, v')$, and thus $f^{\circ}$ is equivalent to a parameterisation of $f$. By Definition~\ref{def:self_similar}, that means that $f^{\circ}$ is self-similar, and thus $f$ is self-similar by equivalence.
\end{proof}
This is an important result as it proves that a \refun with a linear function will produce self-similar functions when connected with other \refuns.

Finally, let us prove that if the function of a \refun is self-similar, then any structure of \refuns is equivalent to having a single \refun.
\bthm
Given two \refuns $r_1$ and $r_2$ both with data type $\mathcal{D}$ (with a set of possible values $D$), with values $v_1\in D$ and $v_2\in D$ respectively, with self-similar function $f: D \times D \to D$, and a connection between them, then the resulting structure is equivalent to having a single \refun.
\ethm
\begin{proof}
    As the function $f$ is self-similar, then for the function $f^{\circ}(i) = f(f(i, v_1), v_2)$ produced by the connection we have that $f^{\circ} \equiv f$ by Definition~\ref{def:self_similar}. As $f^{\circ} \equiv f$ then $\exists v' \in D | f^{\circ}(i) = f(i, v') \forall i \in D$. Then, there exists a \refun $r$ with values $v'$ and function $f$ that is equivalent to the connection between $r_1$ and $r_2$. Thus, the connection is equivalent to have a single \refun.
\end{proof}
This result is critical, because it proves that, for any complex behaviour to emerge from simple functions, it is necessary that those functions are not self-similar, and thus non linear by Theorem~\ref{thm:self-similar}. It also proves that any connection between \refuns with self-similar functions is as powerful as a single \refun can be. This is in turn why Artificial Neural Networks use non linear activation functions to produce emergence~\cite{wb18, krvkhw21}.

Getting back to \refuns with not self-similar functions, let us explore how we can build a setup in which the function is the one that leads the knowledge retrieval and processing. To that end, we need a function able to identify the input and associate it to the required knowledge for processing. However, there is a difficulty here: the emerging function should not only be the knowledge retrieval one, but also the processing one. That is, the function should produce, at the same time, the effect of a knowledge retrieval function that decides which knowledge is necessary to solve the task at hand, and also the effect of the processing function that would take that knowledge and process it to solve the mentioned task. Nevertheless, we have the advantage that, as proven in Theorem~\ref{thm:know}, each one of the \refuns contains knowledge themselves. Thus, we can consider that the basic act of using them to process the input would solve the knowledge retrieval part of the task, as proven in Lemma~\ref{lem:know}. Thus, that leaves us with the task of defining a \refun whose function produces not only emergence, but also the effect of solving the task at hand.

The fact that the \refuns are knowledge by themselves is important because that allows us to store knowledge in multiple ways, both in explicit and implicit forms. This is one of the keys why Artificial Neural Networks manage to obtain so good results: because they store the knowledge implicitly, and use it to transform the input data into the desired output. That hampers explainability, but it has allowed them to solve multiple different problems. A downside of Artificial Neural Networks is that its function is oriented towards producing an optimisation emergent function. That is, its function focuses on getting for the given inputs the optimal output, producing an optimisation effect useful to solve narrow problems but unable to produce the effects of a Cognitive Architecture. That is in part the reason they have so many problems, like catastrophic forgetting\cite{french99, kmahk18} or hallucinations~\cite{okdg+21,jlfy+23}. Thus, novel proposals of \refuns are required, specially ones focused on reproducing the behaviour of a Cognitive Architecture, like the one presented at~\cite{irga24}. However, as this is a theoretical work, the development of such proposals would be matter of future work.

\section{The Function}\label{sec:func}
A fundamental element of any \refun is its function, as it will limit and bound its capabilities. In this sense, a very limiting function will allow to develop systems able to solve a specific kind of problems, but not general enough to solve any other problem. For example, if the function is a simple linear function, then it will be able to solve problems whose solution space can be divided by a linear function. To generalise this notion, the fact is that the chosen function acts like a generator, for which depending on the function the generated function space is different. This function space will be the space of all the functions that the \refun can represent, thus limiting the problems it can solve. If the connection of functions generates new functions through emergence, it can overcome the limitations of the generated function space by increasing it with new generators, but this requires a very specific kind of functions.

Regarding the types of base functions we can produce, we propose a difference between reversible and non-reversible functions, what we call \emph{associative} or \emph{additive} functions respectively. Associative functions will be those that keep intact the associations between the inputs, and thus are able to retrieve those associations also from the output. To fulfil this description, the function needs to be bijective, as we will prove later. In that sense, the function is respecting the association between the input data, hence the name associative. If the function is not bijective, then we have a function that will comprise the initial data into the output without option for reversibility. In that sense, this kind of functions \emph{adds} the input data to produce the output, not respecting the association between the data, hence the name additive. More formally, we need to define additive and associative function.

\bdfn
Given a data type $\mathcal{D}$ with a set of possible values $D$, and a function $f: D \times D \to D$, then $f$ is associative if and only if it keeps the associativity between the input values in the output, that is, the input values are transformed into reversible output values.
\edfn

\bdfn
Given a data type $\mathcal{D}$ with a set of possible values $D$, and a function $f: D \times D \to D$, then $f$ is additive if and only if it collides the values of the input into non-reversible output values.
\edfn

This two definitions will allow us to differentiate between different kinds of functions for our \refuns, but first, let us define them in mathematical terms.

\bthm\label{thm:assoc}
Given a data type $\mathcal{D}$ with a set of possible values $D$, and a function $f: D \times D \to D$, then $f$ is an associative function if and only if it is bijective.
\ethm
\begin{proof}
    Let us start by proving that if $f$ is bijective, then $f$ is associative. Let us assume that $f$ is not associative, then there exists an input value $i$ whose output value $o = f(i)$ is non-reversible, that is, there is no $f_r$ such that $f_r(o) = i$. Now, as $f$ is bijective, then it is injective and surjective. As it is injective, there exists $f_r = f^{-1}$ such that $f_r(o) = f_r(f(i)) = i \forall i \in D$, and thus $f$ should be reversible.

    Let us now prove that if $f$ is associative, then $f$ is bijective. Let us assume that $f$ is not bijective, then it is neither injective or not surjective. Now, as it is associative, it is reversible, meaning that there exists $f_r = f^{-1}$ such that $f_r(o) = f_r(f(i)) = i \forall i \in D$. Thus $f$ is injective. Now, it also is surjective due to the fact that any output value is the image of an input value, thus $f$ is bijective.
\end{proof}

\blem
Given a data type $\mathcal{D}$ with a set of possible values $D$, and a function $f: D \times D \to D$, then $f$ is an additive function if and only if it is not bijective.
\elem
\begin{proof}
    As shown in the proof of Theorem~\ref{thm:assoc}, if $f$ is not bijective, then it collides the input values into non-reversible output values, thus it is additive.
\end{proof}
This result is important because it shows that Artificial Neural Networks use an additive function in their \refuns. This will have implications in the kind of functions they can represent. It is important to notice that, given the context in which we are working, it is very rare to find a function that is not surjective, but for the sake of completeness, we will consider that case too.

As an additional theorem, let us prove that an additive function produces a loss of information.
\bthm\label{thm:info_loss}
Given a data type $\mathcal{D}$ with a set of possible values $D$, and a function $f: D \times D \to D$ that is an additive function, then $f$ loses information.
\ethm
\begin{proof}
    Considering Shannon's definition of information as entropy of a set, we have that a smaller set has less entropy and thus less information than a larger set. Now, as $f$ is an additive function, it is not bijective, either because it is not surjective or because it is not injective. If $f$ is not surjective, its output set $O$ does not cover all the output space $D$, thus $O \subsetneq D$. If $f$ is not injective, as the input set $I = D$, by the Pigeonhole Principle there are at least two inputs that collide into the same output, and that implies that $O \subsetneq D$. In either case, that implies that the output set $O$ is smaller than the input set $I$, and thus there is a loss of information after applying $f$.
\end{proof}
This result will be useful in the following proofs, as it is the cornerstone over which the difference between additive and associative functions revolve.

Now, let us prove that a \refun with additive function is limited in the type of functions it can perform.
\bthm
Given a data type $\mathcal{D}$ with a set of possible values $D$, and a function $f: D \times D \to D$ that is an additive function, then $f$ cannot be equivalent to a bijective function.
\ethm
\begin{proof}
    Let us take, without loss of generality, the bijective function \emph{par excellence}: the identity function. If $f$ is an additive function, it is not bijective. Thus, it either collides multiple inputs into the same output (it is not injective) or it does not cover the complete output space (it is not surjective). In either case, there is at least an input $i \in D$ whose output $f(i) = o \in D$ is not itself ($i \neq o$). Thus, it is impossible to represent the identity function. Moreover, a loss of information is produced, as the output set has less elements than the input set.
    
    If we consider a connection between multiple \refuns, we have that, once we have lost the information of the original input in the outputs of the first \refun, the subsequent \refuns cannot recover that information, as they are also additive, and thus, by Theorem~\ref{thm:info_loss}, they keep losing information, instead of recovering it.
\end{proof}
This result shows that any \refun with additive function has a limitation in the kind of functions it can represent. Specifically, it can not represent bijective functions, and thus it can not perform a basic task of Cognitive Architectures as associating different concepts all by itself. However, it is important to remark that this limitation can be overcome through emergence and specifically built architectures. For example, Artificial Neural Networks overcome this limitation by connecting multiple \refuns with non linear functions, being able to build architectures like Variational Autoencoders~\cite{dmglsas16} whose whole goal is to recreate the identity function.

Symmetrically, we can also prove that associative functions do not produce a loss of information and thus they also have limits in their capability to produce functions.
\bthm\label{thm:no_info}
Given a data type $\mathcal{D}$ with a set of possible values $D$, and a function $f: D \times D \to D$ that is an associative function, then $f$ does not lose information.
\ethm
\begin{proof}
    Considering Shannon's definition of information as entropy of a set, we have that a smaller set has less entropy and thus less information than a larger set. Now, as $f$ is an associative function, it is bijective. As $f$ is surjective, its output set $O$ covers all the output space $D$, thus $O = D$. As $f$ is injective and the input set $I = D$, there are no two inputs that collide into the same output, that implies that $O = D$. Thus, the output set $O$ is equal to the input set $I$, and thus there is no loss of information.
\end{proof}
\bthm
Given a data type $\mathcal{D}$ with a set of possible values $D$, and a function $f: D \times D \to D$ that is an associative function, then $f$ cannot be equivalent to a non-bijective function.
\ethm
\begin{proof}
    Let us take, without loss of generality, the non-bijective function \emph{par excellence}: the constant function. If $f$ is an associative function, it is bijective. Thus, it does not collide multiple inputs into the same output (it is injective) and it covers the complete output space (it is surjective). Thus, there is no pair of inputs $i_1, i_2 \in D$ whose outputs $f(i_1) = o_1 \in D$ and $f(i_2) = o_2 \in D$ are the same ($o_1 \neq o_2$). Thus, it is impossible to represent the constant function. Moreover, no loss of information is produced, as the output set has the same amount of elements than the input set.
    
    If we consider a connection between multiple \refuns, we have that, as we do not lose the information of the original input in the outputs of the first \refun, and subsequent \refuns do not lose that information either, as they are also associative, then, by Theorem~\ref{thm:no_info}, they keep not losing information.
\end{proof}
This result shows that any \refun with associative function has a limitation in the kind of functions it can represent. Specifically, it can not represent non-bijective functions, and thus it can not perform a basic task of Cognitive Architectures as abstracting different elements into a concept. However, similarly as before, this limitation may be overcome through emergence and specifically built architectures.

\section{Case Studies}\label{sec:cases}
In this Section we will introduce some case studies that show how this theory can be implemented. Specifically, we will explore the different cases based on how the \refuns are connected. We will start with the sequential case, in which we will use Artificial Neural Networks as an example. We will continue with the parallel case, where we present a very basic proposal of \refun. And we will finish with the hybrid case between sequential and parallel, where we will present a more complex proposal of \refun. It is important to note that this section will be more of a thought experiment, where we will try to base our thoughts in the theory outlined in the previous sections.

\subsection{The Sequential Case}
Let us explore first the sequential case. This case is defined by organising a set of \refuns connecting one after another in a sequential fashion. As an example of a sequential \refun we will explore Artificial Neural Networks. They are a well known implementation of \refun where each neuron is an individual \refun whose representation are the weights and whose function is comprised of three steps: the multiplication of each weight for the value of the input, the sum of all the resulting values, and the application of a non linear function to the result. As it can be derived from this definition, in the end each neuron is a parameterised function, whose parameters are the internal representation of the acquired knowledge. In the case of neurons, the \refun aims to build any kind of mapping function from a space of inputs to a space of outputs. Thus, the set of inputs typically has a different dimension size than the set of outputs and the training focuses on minimising the error between the produced output and the expected output. This setup is more suited for optimisation and generalisation, although not so useful for building a Cognitive Architecture.

The fact that the dimension size of the input and output of a neuron is not the same allows the neurons to be placed in parallel to process differently the same input, and thus being able to provide an input of dimension higher than $1$ to the next layer of neurons. However, this does not make the Artificial Neural Networks a sample of the parallel case, because in the end all this structure works like a sequential set of \refuns, as the output is unique for each input.

\subsection{The Parallel Case}
Let us explore now the parallel case. In this case, the determining factor is that multiple \refuns process the same input in parallel and produce multiple outputs, from which later we have to decide which one to keep. To propose a \refun suited for this case, we will explore a basic algorithm, maybe more suited for Cognitive Architectures. This algorithm will be an identification algorithm, where the \refun contains a given representation, and decides if the input is recognised as equivalent to such representation using a similarity function and a threshold. The output will be either the representation (if the similarity between the input and the representation was over the threshold) or nothing. This setup would work as a memory, where each \refun would be a sample and the result of the whole system would be the identification from memory of the input.

\subsection{The Hybrid Case}
Finally, let us explore a hybrid case between the parallel and sequential ones. In this case, the defining factor is that we will have multiple \refuns that would process the same input, but at the same time there will be another set of \refuns, connected to the first ones, that would provide depth to the structure. To propose a \refun suited for this case, and following the idea of the parallel case, we propose the previous \refun applying a final non linear function to its output. In this case, thanks to the non linearity of this \refun, we can connect multiple \refuns one after another in order to build more complex identification functions. Additionally, we can also build multiple non-connected \refuns and choose between them based on their similarity to the input, with the additional depth allowing the identification of more complex inputs.

\section{Limitations}\label{sec:disc}
In this section we will address the main limitation of this work, that is its pure theoretical approach. Although we have used Artificial Neural Networks as example and reference during the paper, we have not empirically validated that the implementation of this theory could develop a Cognitive Architecture, mainly due to the fact that building a whole Cognitive Architecture with this approach falls outside the scope of this paper. However, an outline of how this theory can be used to produce a Cognitive Architecture is presented at~\cite{irga24}. We are aware that sometimes the limitation of a theory is its impossibility to be applied, and we tried to address this issue presenting some case studies in Section~\ref{sec:cases}, but in the end those are thought experiments rather than an empirical validation. In the case of Artificial Neural Networks, we have supported our claims with cites to the already existing literature. However, we are aware that they are not considered a proper Cognitive Architecture but rather an optimisation algorithm, and thus they are not enough to claim an empirical validation of this theory, although it is still an empirical validation of this theory with regard to the characteristics of an individual \refun or a chain of connected \refuns.

\section{Conclusions}\label{sec:conc}
Traditional and widely known Cognitive Architectures have a huge problem with knowledge retrieval, as the heuristics to know which elements to retrieve from memory are very complex, maybe even impossible to achieve. To address this problem, in this paper we proposed to change the fundamental model of computation to the \refun one, where memory and program are unified into a single entity. This model can solve the problem of knowledge retrieval through the dynamic adaptation of the program to the memory, deprecating the need of a knowledge retrieval heuristic.

To back our proposal, we have presented some definitions and theorems that justify our claims, and proven them with mathematical proofs. Additionally, we have carried out thought experiments to show how our theory can be applied to build different \refuns, although the development of a proper Cognitive Architecture based on a \refun is a matter of future work. During the paper we have used Artificial Neural Networks to inspire and guide our claims and proofs, and as example to facilitate the understanding of concepts. Finally, we have explored some limitations of our proposal due to its pure theoretical nature.

As future work, we would like to implement this theory to build a proper Cognitive Architecture based on a \refun, and we would like to empirically validate our claims, in particular those related with Cognitive Architectures. We would also like to explore new proposals of \refuns that could solve certain problems, even if they do not build a Cognitive Architecture. Finally, we would like to explore new properties of a \refun that could lead to new results and properties.

\section*{Acknowledgements}
We want to thank Daniel Pinyol and Pere Mayol for our insightful discussions about the topic.

\bibliographystyle{IEEEtran}
\bibliography{biblio}

\begin{thebibliography}{10}
\providecommand{\url}[1]{#1}
\csname url@samestyle\endcsname
\providecommand{\newblock}{\relax}
\providecommand{\bibinfo}[2]{#2}
\providecommand{\BIBentrySTDinterwordspacing}{\spaceskip=0pt\relax}
\providecommand{\BIBentryALTinterwordstretchfactor}{4}
\providecommand{\BIBentryALTinterwordspacing}{\spaceskip=\fontdimen2\font plus
\BIBentryALTinterwordstretchfactor\fontdimen3\font minus \fontdimen4\font\relax}
\providecommand{\BIBforeignlanguage}[2]{{%
\expandafter\ifx\csname l@#1\endcsname\relax
\typeout{** WARNING: IEEEtran.bst: No hyphenation pattern has been}%
\typeout{** loaded for the language `#1'. Using the pattern for}%
\typeout{** the default language instead.}%
\else
\language=\csname l@#1\endcsname
\fi
#2}}
\providecommand{\BIBdecl}{\relax}
\BIBdecl

\bibitem{Laird08}
J.~E. Laird, ``Extending the soar cognitive architecture,'' in \emph{Artificial General Intelligence 2008, Proceedings of the First {AGI} Conference, {AGI} 2008, March 1-3, 2008, University of Memphis, Memphis, TN, {USA}}, ser. Frontiers in Artificial Intelligence and Applications, vol. 171.\hskip 1em plus 0.5em minus 0.4em\relax {IOS} Press, 2008, pp. 224--235.

\bibitem{Laird12}
------, \emph{The Soar Cognitive Architecture}.\hskip 1em plus 0.5em minus 0.4em\relax The MIT Press, 2012.

\bibitem{VanLehn86}
K.~VanLehn, ``John r. anderson, the architecture of cognition,'' \emph{Artif. Intell.}, vol.~28, no.~2, pp. 235--240, 1986.

\bibitem{Mitchell90}
M.~Mitchell, ``Copycat: {A} computer model of high-level perception and conceptual slippage in analogy making,'' Ph.D. dissertation, University of Michigan, {USA}, 1990.

\bibitem{Burns95}
B.~D. Burns, ``Fluid concepts and creative analogies: {A} review,'' \emph{{AI} Mag.}, vol.~16, no.~3, pp. 81--83, 1995.

\bibitem{Lieto16}
A.~Lieto, ``Representational limits in cognitive architectures,'' in \emph{Cognitive Robot Architectures, Proceedings of EUCognition 2016 European Association for Cognitive Systems, Vienna, Austria, 8-9 December, 2016}, ser. {CEUR} Workshop Proceedings, vol. 1855.\hskip 1em plus 0.5em minus 0.4em\relax CEUR-WS.org, 2016, pp. 16--20.

\bibitem{pc17}
D.~Peebles and P.~C. Cheng, ``Multiple representations in cognitive architectures,'' in \emph{2017 {AAAI} Fall Symposia, Arlington, Virginia, USA, November 9-11, 2017}.\hskip 1em plus 0.5em minus 0.4em\relax {AAAI} Press, 2017, pp. 425--430.

\bibitem{ii24}
A.~Ismayilova and V.~E. Ismailov, ``On the kolmogorov neural networks,'' \emph{Neural Networks}, vol. 176, p. 106333, 2024.

\bibitem{yww18}
P.~Ye, T.~Wang, and F.~Wang, ``A survey of cognitive architectures in the past 20 years,'' \emph{{IEEE} Trans. Cybern.}, vol.~48, no.~12, pp. 3280--3290, 2018.

\bibitem{Albus00}
J.~S. Albus, ``4-d/rcs reference model architecture for unmanned ground vehicles,'' in \emph{Proceedings of the 2000 {IEEE} International Conference on Robotics and Automation, {ICRA} 2000, April 24-28, 2000, San Francisco, CA, {USA}}.\hskip 1em plus 0.5em minus 0.4em\relax {IEEE}, 2000, pp. 3260--3265.

\bibitem{abchss06}
J.~S. Albus, R.~Bostelman, T.~Chang, T.~Hong, W.~P. Shackleford, and M.~Shneier, ``Learning in a hierarchical control system: 4d/rcs in the {DARPA} {LAGR} program,'' \emph{J. Field Robotics}, vol.~23, no. 11-12, pp. 975--1003, 2006.

\bibitem{bll04}
D.~P. Benjamin, D.~M. Lyons, and D.~W. Lonsdale, ``{ADAPT:} {A} cognitive architecture for robotics,'' in \emph{Proceedings of the International Conference on Cognitive Modelling, {ICCM} 2004, Pittsburgh, Pennsylvania, USA, July 30 - August 1, 2004}, 2004, pp. 337--338.

\bibitem{bll04b}
D.~P. Benjamin, D.~Lonsdale, and D.~M. Lyons, ``Designing a robot cognitive architecture with concurrency and active perception,'' in \emph{The Intersection of Cognitive Science and Robotics: From Interfaces to Intelligence, Papers from the 2004 {AAAI} Fall Symposium. Arlington, VA, USA, October 22-24, 2004}, vol. {FS-04-05}.\hskip 1em plus 0.5em minus 0.4em\relax {AAAI} Press, 2004, pp. 1--8.

\bibitem{haht11}
T.~Huntsberger, H.~Aghazarian, A.~Howard, and D.~C. Trotz, ``Stereo vision-based navigation for autonomous surface vessels,'' \emph{J. Field Robotics}, vol.~28, no.~1, pp. 3--18, 2011.

\bibitem{mmadfjj+01}
B.~A. Maxwell, L.~Meeden, N.~S. Addo, P.~Dickson, N.~Fairfield, N.~Johnson, E.~G. Jones, S.~Kim, P.~Malla, M.~Murphy, B.~Rutter, and E.~Silk, ``Reaper: {A} reflexive architecture for perceptive agents,'' \emph{{AI} Mag.}, vol.~22, no.~1, pp. 53--66, 2001.

\bibitem{pttan12}
X.~Pan, L.~Teow, K.~H. Tan, J.~H.~B. Ang, and G.~W. Ng, ``A cognitive system for adaptive decision making,'' in \emph{15th International Conference on Information Fusion, {FUSION} 2012, Singapore, July 9-12, 2012}.\hskip 1em plus 0.5em minus 0.4em\relax {IEEE}, 2012, pp. 1323--1329.

\bibitem{nxct12}
G.~W. Ng, X.~H. Xiao, R.~Z. Chan, and Y.~Tan, ``Scene understanding using {DSO} cognitive architecture,'' in \emph{15th International Conference on Information Fusion, {FUSION} 2012, Singapore, July 9-12, 2012}.\hskip 1em plus 0.5em minus 0.4em\relax {IEEE}, 2012, pp. 2277--2284.

\bibitem{Wang13}
P.~Wang, ``Natural language processing by reasoning and learning,'' in \emph{Artificial General Intelligence - 6th International Conference, {AGI} 2013, Beijing, China, July 31 - August 3, 2013 Proceedings}, ser. Lecture Notes in Computer Science, vol. 7999.\hskip 1em plus 0.5em minus 0.4em\relax Springer, 2013, pp. 160--169.

\bibitem{swxw15}
N.~Slam, W.~Wang, G.~Xue, and P.~Wang, ``A framework with reasoning capabilities for crisis response decision-support systems,'' \emph{Eng. Appl. Artif. Intell.}, vol.~46, pp. 346--353, 2015.

\bibitem{skms06}
P.~W. Schermerhorn, J.~F. Kramer, C.~Middendorff, and M.~Scheutz, ``{DIARC:} {A} testbed for natural human-robot interaction,'' in \emph{Proceedings, The Twenty-First National Conference on Artificial Intelligence and the Eighteenth Innovative Applications of Artificial Intelligence Conference, July 16-20, 2006, Boston, Massachusetts, {USA}}.\hskip 1em plus 0.5em minus 0.4em\relax {AAAI} Press, 2006, pp. 1972--1973.

\bibitem{gpptw98}
M.~P. Georgeff, B.~Pell, M.~E. Pollack, M.~Tambe, and M.~J. Wooldridge, ``The belief-desire-intention model of agency,'' in \emph{Intelligent Agents V, Agent Theories, Architectures, and Languages, 5th International Workshop, {ATAL} '98, Paris, France, July 4-7, 1998, Proceedings}, ser. Lecture Notes in Computer Science, vol. 1555.\hskip 1em plus 0.5em minus 0.4em\relax Springer, 1998, pp. 1--10.

\bibitem{mg01}
J.~W. Murdock and A.~K. Goel, ``Meta-case-based reasoning: Using functional models to adapt case-based agents,'' in \emph{Case-Based Reasoning Research and Development, 4th International Conference on Case-Based Reasoning, {ICCBR} 2001, Vancouver, BC, Canada, July 30 - August 2, 2001, Proceedings}, ser. Lecture Notes in Computer Science, vol. 2080.\hskip 1em plus 0.5em minus 0.4em\relax Springer, 2001, pp. 407--421.

\bibitem{njw10}
R.~Novianto, B.~Johnston, and M.~Williams, ``Attention in the {ASMO} cognitive architecture,'' in \emph{Biologically Inspired Cognitive Architectures 2010 - Proceedings of the First Annual Meeting of the {BICA} Society, Washington, DC, USA, November 13-14, 2010}, ser. Frontiers in Artificial Intelligence and Applications, vol. 221.\hskip 1em plus 0.5em minus 0.4em\relax {IOS} Press, 2010, pp. 98--105.

\bibitem{mbv12}
Z.~Mathews, S.~B. i~Badia, and P.~F. M.~J. Verschure, ``{PASAR:} an integrated model of prediction, anticipation, sensation, attention and response for artificial sensorimotor systems,'' \emph{Inf. Sci.}, vol. 186, no.~1, pp. 1--19, 2012.

\bibitem{bfpd06}
F.~Bellas, A.~Fai{\~{n}}a, A.~Prieto, and R.~J. Duro, ``Adaptive learning application of the {MDB} evolutionary cognitive architecture in physical agents,'' in \emph{From Animals to Animats 9, 9th International Conference on Simulation of Adaptive Behavior, {SAB} 2006, Rome, Italy, September 25-29, 2006, Proceedings}, ser. Lecture Notes in Computer Science, vol. 4095.\hskip 1em plus 0.5em minus 0.4em\relax Springer, 2006, pp. 434--445.

\bibitem{bb15}
W.~Bridewell and P.~Bello, ``Incremental object perception in an attention-driven cognitive architecture,'' in \emph{Proceedings of the 37th Annual Meeting of the Cognitive Science Society, CogSci 2015, Pasadena, California, USA, July 22-25, 2015}.\hskip 1em plus 0.5em minus 0.4em\relax cognitivesciencesociety.org, 2015.

\bibitem{fmsfd+16}
S.~Franklin, T.~Madl, S.~Strain, U.~Faghihi, D.~Dong, S.~Kugele, J.~Snaider, P.~Agrawal, and S.~Chen, ``A lida cognitive model tutorial,'' \emph{Biologically Inspired Cognitive Architectures}, vol.~16, pp. 105--130, 2016.

\bibitem{js13}
M.~Jaszuk and J.~A. Starzyk, ``Building internal scene representation in cognitive agents,'' in \emph{Knowledge, Information and Creativity Support Systems: Recent Trends, Advances and Solutions - Selected Papers from KICSS'2013 - 8th International Conference on Knowledge, Information, and Creativity Support Systems, November 7-9, 2013, Krak{\'{o}}w, Poland}, ser. Advances in Intelligent Systems and Computing, vol. 364.\hskip 1em plus 0.5em minus 0.4em\relax Springer, 2013, pp. 479--491.

\bibitem{edelman07}
G.~M. Edelman, ``Learning in and from brain-based devices,'' \emph{science}, vol. 318, no. 5853, pp. 1103--1105, 2007.

\bibitem{ha15}
\BIBentryALTinterwordspacing
J.~Hawkins and S.~Ahmad, ``Why neurons have thousands of synapses, {A} theory of sequence memory in neocortex,'' \emph{CoRR}, vol. abs/1511.00083, 2015. [Online]. Available: \url{http://arxiv.org/abs/1511.00083}
\BIBentrySTDinterwordspacing

\bibitem{lklg14}
M.~Lloyd{-}Kelly, P.~C.~R. Lane, and F.~Gobet, ``The effects of bounding rationality on the performance and learning of {CHREST} agents in tileworld,'' in \emph{Research and Development in Intelligent Systems XXXI, Incorporating Applications and Innovations in Intelligent Systems {XXII.} Proceedings of AI-2014, The Thirty-fourth {SGAI} International Conference on Innovative Techniques and Applications of Artificial Intelligence, Cambridge, UK, December 9-11, 2014}.\hskip 1em plus 0.5em minus 0.4em\relax Springer, 2014, pp. 149--162.

\bibitem{spm99}
R.~Sun, T.~Peterson, and E.~Merrill, ``A hybrid architecture for situated learning of reactive sequential decision making,'' \emph{Appl. Intell.}, vol.~11, no.~1, pp. 109--127, 1999.

\bibitem{jloa08}
D.~J. Jilk, C.~Lebiere, R.~C. O'Reilly, and J.~R. Anderson, ``{SAL:} an explicitly pluralistic cognitive architecture,'' \emph{J. Exp. Theor. Artif. Intell.}, vol.~20, no.~3, pp. 197--218, 2008.

\bibitem{llr17}
J.~E. Laird, C.~Lebiere, and P.~S. Rosenbloom, ``A standard model of the mind: Toward a common computational framework across artificial intelligence, cognitive science, neuroscience, and robotics,'' \emph{{AI} Mag.}, vol.~38, no.~4, pp. 13--26, 2017.

\bibitem{gbkgm19}
S.~Ghose, A.~Boroumand, J.~S. Kim, J.~G{\'{o}}mez{-}Luna, and O.~Mutlu, ``Processing-in-memory: {A} workload-driven perspective,'' \emph{{IBM} J. Res. Dev.}, vol.~63, no.~6, pp. 3:1--3:19, 2019.

\bibitem{klr19}
Y.~Kwon, Y.~Lee, and M.~Rhu, ``Tensordimm: {A} practical near-memory processing architecture for embeddings and tensor operations in deep learning,'' in \emph{Proceedings of the 52nd Annual {IEEE/ACM} International Symposium on Microarchitecture, {MICRO} 2019, Columbus, OH, USA, October 12-16, 2019}.\hskip 1em plus 0.5em minus 0.4em\relax {ACM}, 2019, pp. 740--753.

\bibitem{rkr16}
A.~Rahimi, P.~Kanerva, and J.~M. Rabaey, ``A robust and energy-efficient classifier using brain-inspired hyperdimensional computing,'' in \emph{Proceedings of the 2016 International Symposium on Low Power Electronics and Design, {ISLPED} 2016, San Francisco Airport, CA, USA, August 08 - 10, 2016}.\hskip 1em plus 0.5em minus 0.4em\relax {ACM}, 2016, pp. 64--69.

\bibitem{sha48}
C.~E. Shannon, ``A mathematical theory of communication,'' \emph{The Bell System Technical Journal}, vol.~27, pp. 379--423, 623--656, 1948.

\bibitem{wb18}
T.~Wiatowski and H.~B{\"{o}}lcskei, ``A mathematical theory of deep convolutional neural networks for feature extraction,'' \emph{{IEEE} Trans. Inf. Theory}, vol.~64, no.~3, pp. 1845--1866, 2018.

\bibitem{krvkhw21}
N.~Kulathunga, N.~R. Ranasinghe, D.~Vrinceanu, Z.~Kinsman, L.~Huang, and Y.~Wang, ``Effects of nonlinearity and network architecture on the performance of supervised neural networks,'' \emph{Algorithms}, vol.~14, no.~2, p.~51, 2021.

\bibitem{french99}
R.~M. French, ``Catastrophic forgetting in connectionist networks,'' \emph{Trends in cognitive sciences}, vol.~3, no.~4, pp. 128--135, 1999.

\bibitem{kmahk18}
R.~Kemker, M.~McClure, A.~Abitino, T.~Hayes, and C.~Kanan, ``Measuring catastrophic forgetting in neural networks,'' in \emph{Proceedings of the AAAI conference on artificial intelligence}, vol.~32, no.~1, 2018.

\bibitem{okdg+21}
\BIBentryALTinterwordspacing
P.~A. Ortega, M.~Kunesch, G.~Del{\'{e}}tang, T.~Genewein, J.~Grau{-}Moya, J.~Veness, J.~Buchli, J.~Degrave, B.~Piot, J.~P{\'{e}}rolat, T.~Everitt, C.~Tallec, E.~Parisotto, T.~Erez, Y.~Chen, S.~E. Reed, M.~Hutter, N.~de~Freitas, and S.~Legg, ``Shaking the foundations: delusions in sequence models for interaction and control,'' \emph{CoRR}, vol. abs/2110.10819, 2021. [Online]. Available: \url{https://arxiv.org/abs/2110.10819}
\BIBentrySTDinterwordspacing

\bibitem{jlfy+23}
Z.~Ji, N.~Lee, R.~Frieske, T.~Yu, D.~Su, Y.~Xu, E.~Ishii, Y.~J. Bang, A.~Madotto, and P.~Fung, ``Survey of hallucination in natural language generation,'' \emph{ACM Comput. Surv.}, vol.~55, no.~12, 2023.

\bibitem{irga24}
A.~Ibias, G.~Ramirez{-}Miranda, E.~Guinovart, and E.~Alarc{\'{o}}n, ``From manifestations to cognitive architectures: {A} scalable framework,'' in \emph{Artificial General Intelligence - 17th International Conference, {AGI} 2024, Seattle, WA, USA, August 13-16, 2024, Proceedings}, ser. Lecture Notes in Computer Science, vol. 14951.\hskip 1em plus 0.5em minus 0.4em\relax Springer, 2024, pp. 89--98.

\bibitem{dmglsas16}
\BIBentryALTinterwordspacing
N.~Dilokthanakul, P.~A.~M. Mediano, M.~Garnelo, M.~C.~H. Lee, H.~Salimbeni, K.~Arulkumaran, and M.~Shanahan, ``Deep unsupervised clustering with gaussian mixture variational autoencoders,'' \emph{CoRR}, vol. abs/1611.02648, 2016. [Online]. Available: \url{http://arxiv.org/abs/1611.02648}
\BIBentrySTDinterwordspacing

\end{thebibliography}


\begin{IEEEbiographynophoto}{Alfredo Ibias}
received B.Sc. degrees in Computer Science and in Mathematics from Complutense University of Madrid, Spain, and an M.Sc. degree in Formal Methods in Computer Science and a Ph.D. degree in Computer Science from the same university. For 4 years he did research around the development of AI methods for uncommon scenarios. Currently he is working as an AI researcher at Avatar Cognition. His main research area is the development of a general AI based on novel theories of the brain.
\end{IEEEbiographynophoto}

\begin{IEEEbiographynophoto}{Hector Antona}
received B.Sc. degrees in Computer Science and in Telecommunications Engineering from Universitat Politècnica de Catalunya (UPC), Spain, and an M.Sc. degree in Advanced Telecomunications Technologies from the same university. Currently he is working as an AI researcher at Avatar Cognition. His main research area is the applicability of novel AI methods.
\end{IEEEbiographynophoto}

\begin{IEEEbiographynophoto}{Guillem Ramirez-Mirandaz}
received B.Sc. degree in Computer Science from Universitat Politécnica de Barcelona, Spain. For 3 years he did research on performance analysis and optimisation of high-performance computing applications. Currently, he is working as a developer and researcher at Avatar Cognition and pursuing a B.A. degree in Philosophy at Universidad Nacional de Estudios a Distancia (UNED), Spain. His main research area is the development of a general AI based on novel theories of the brain.
\end{IEEEbiographynophoto}

\begin{IEEEbiographynophoto}{Enric Guinovart}
received B.Sc. degree in Computer Science from Universitat Politècnica de Catalunya (UPC), Spain. He has been working in the industry for 20 years as AI consultant (among other roles). In 2018 he funded Avatar Cognition, where he currently works as co-CEO, CTO and CRO. His main research area is the development of a general AI based on novel theories of the brain.
\end{IEEEbiographynophoto}

\begin{IEEEbiographynophoto}{Eduard Alarcon} 
received the M.Sc. (National award) and Ph.D. degrees (honors) in Electrical Engineering from the Technical University of Catalunya (UPC BarcelonaTech), Spain, in 1995 and 2000, respectively. Since 1995 he has been with the Department of Electronics Engineering at the School of Telecommunications at UPC, where he became Associate Professor in 2000 and is currently full professor. Visiting professor at CU Boulder and KTH. He has co-authored more than 450 scientific publications, 6 books, 8 book chapters and 12 patents. He has been involved in different National, European (H2020 FET-Open, Flag-ERA, ESA) and US (DARPA, NSF, NASA) R\&D projects within his research interests including the areas of on-chip energy management and RF circuits, energy harvesting and wireless energy transfer, nanosatellites and satellite architectures for Earth Observation, nanotechnology-enabled wireless communications, Quantum computing architectures and Artificial Intelligence chip architectures. He has received the GOOGLE Faculty Research Award (2013), SAMSUNG Advanced Institute of Technology Global Research Program gift (2012), and INTEL Doctoral Student Honor Programme Fellowship (2014). Professional officer responsibilities include elected member of the IEEE CAS Board of Governors (2010-2013) and Vice President for Technical Activities of IEEE CAS (2016-2017, and 2017-2018). Editorial duties include Senior founding Editorial Board of the IEEE Journal on Emerging topics in Circuits and Systems, of which he was Editor-in-Chief (2018-2019).
\end{IEEEbiographynophoto}

\vfill

\end{document}